\documentclass[10pt]{article}

\usepackage[margin=0.82in]{geometry}
\usepackage{amsmath,amssymb}
\usepackage{booktabs}
\usepackage{graphicx}
\usepackage{microtype}
\usepackage{natbib}
\usepackage{caption}
\usepackage{subcaption}
\usepackage{enumitem}
\usepackage[hidelinks]{hyperref}
\hypersetup{pdftitle={Knowledge as Orbit: Finite Collections as Phases of an Exactly Periodic Latent Generator},pdfauthor={Siddharth Pal and Viktoria Rojkova}}
\usepackage[T1]{fontenc}
\usepackage{lmodern}

\newcommand{\R}{\mathbb{R}}
\newcommand{\Id}{\mathrm{I}}

\title{\textbf{Knowledge as Orbit:}\\
Finite Collections as Phases of an Exactly Periodic Latent Generator}
\author{Siddharth Pal \and Viktoria Rojkova}
\date{Preprint}

\begin{document}
\maketitle

\begin{abstract}
Finite knowledge is usually represented extensionally: one stored code, record, or parameter vector per item. We ask whether a finite collection can instead be represented intensionally, by a compact law whose orbit enumerates the items and returns exactly to its start. For a collection of $X$ objects, we encode item $i$ as the $i$-th phase of one fixed rotation in a learned latent space and decode all phases with a shared nonlinear network. The latent advances through a bank of rotations at integer harmonics of the cycle---a real discrete Fourier operator---so $R^X=\Id$ and exact closure is guaranteed rather than learned. Images provide a controlled carrier for measuring the representation, while looping video is the case in which the imposed phase order is the content's own temporal structure.

The exact periodicity is the mechanism, not a detail. Holding the decoder fixed and varying only the operator, a general learned operator is erratic and eventually diverges, a norm-preserving but non-periodic operator degrades around the loop, and the exactly periodic operator is flat; on real images the gap widens. With closure handled by the operator, capacity is the decoder's budget, and we account for it: dense decoders carry a structural overhead of roughly $2.25\times$ per crisp image that no size reconciles with compression, while a small convolutional decoder on objects that share a manifold reaches crisp and compressed. A codebook control shows that the generative law is free in reconstruction terms: replacing the cycle with an independent learned latent per item never improves reconstruction while multiplying the latent store one-hundred-twenty-eight-fold. On seven benchmark clips, against a matched NeRV-style frame-index baseline, the cycle reaches equal or better fidelity at equal parameters while wrapping at machine precision, where the frame-index representation leaves a two-to-five percent seam; pinning the baseline's frequencies to loop harmonics closes its seam as well, confirming that exact periodicity, however imposed, is the operative constraint. All results are small-scale and internally controlled; compression is measured against raw frames and matched baselines, not against image or video codecs. The resulting claim is deliberately scoped: finite cyclic knowledge can be stored as dynamics rather than as independent instances, with exact recurrence supplied by algebra and content supplied by a shared decoder.
\end{abstract}

\section{Finite knowledge as a generated orbit}

A finite collection is conventionally stored as a list,
\[
\mathcal{X}=\{x_0,x_1,\ldots,x_{X-1}\},
\]
with one address or latent code per item. That representation says what the items are, but it does not provide a law that moves among them. We study the alternative in which the collection is the decoded orbit of one dynamical generator:
\[
x_0 \rightarrow x_1 \rightarrow \cdots \rightarrow x_{X-1} \rightarrow x_0.
\]
The collection is then represented by an initial latent state, a fixed advance operator, and a shared decoder. Access by index becomes phase selection; sequential access becomes repeated application of the same operator; closure becomes an algebraic property of the representation.

We use the term \emph{finite knowledge} operationally: a finite set of objects that a model must reproduce, not a claim about propositional truth or reasoning. The construction requires a cyclic indexing of those objects. For an unordered image set, that order is imposed and serves as a stringent capacity test: the representation must hold unrelated objects without drift, but intermediate phases need not be semantically meaningful. For looping video, the phase order is intrinsic: it is time, and exact closure is a required property of the signal. This distinction---an imposed orbit for finite storage versus a natural orbit for cyclic phenomena---is central to what the experiments do and do not establish.

The question is therefore broader than whether a transform can map one image to another. It is whether a finite collection can be represented by a law rather than by independent instances, and what exact recurrence, capacity, and compression cost that law entails. A random invertible map sends an image to noise because natural images occupy a thin manifold in pixel space. The problem is to keep every decoded phase on the object manifold and to make the latent orbit close. We place periodicity in the latent as a Fourier rotation and object validity in a learned decoder. Images are the controlled measurement domain; looping video is the naturally cyclic application.

\paragraph{Contributions.} We make three scoped claims. First, exact periodicity is the structural constraint that lets a single operator enumerate many objects without accumulated drift, where a general operator diverges and a merely norm-preserving one fails to close. Second, that constraint costs nothing against independent latent codes in the reconstruction settings measured here, while using far less latent storage and supplying a total order and exact closure. Third, when the represented collection is naturally cyclic, as in looping video, the construction converts an approximate training outcome, the loop seam, into a property guaranteed by the representation rather than learned.

\section{Method: one law, $X$ phases}

\subsection{Finite objects as Fourier phases}

We represent a cycle by a fixed rotation $R$ acting on a latent $z\in\R^d$ and a nonlinear decoder $D$ that renders a latent as an object in $\R^N$,
\begin{equation}
 z_i = R^i z_0, \qquad x_i \approx D(z_i), \qquad i=0,\ldots,X-1,
 \label{eq:orbit}
\end{equation}
and train only the initial latent $z_0$ and the decoder $D$ to fit the target collection $\{x_i\}$. For the experiments below, the objects are images or video frames.

The operator is a block-diagonal bank of plane rotations at integer harmonic frequencies $f_1,\ldots,f_m$ of the fundamental $2\pi/X$,
\begin{equation}
R = \operatorname{blockdiag}\!\left(\rho(2\pi f_j/X)\right)_{j=1}^{m},
\qquad
\rho(\theta)=
\begin{pmatrix}
\cos\theta & -\sin\theta\\
\sin\theta & \cos\theta
\end{pmatrix}.
\label{eq:fourier}
\end{equation}
This is the real discrete Fourier operator: the $j$-th plane is a phasor spinning at a harmonic of the cycle. Because the frequencies are integers, $R^X=\Id$ and the orbit closes exactly. The decoded sequence $D(z_0),D(z_1),\ldots$ is a learned Fourier series in the item index. The rotation is both an address generator and a clock; the decoder is the renderer that assigns content to each address.

Viewed as a finite knowledge representation, Equation~\eqref{eq:orbit} replaces $X$ independent latent addresses with one base state and a known group action of the cyclic group $C_X$. It is memory through dynamics: the orbit specifies where each item is, while the decoder specifies what each item is.

\subsection{Why exact periodicity matters}

Nothing forbids a general learned latent operator, as in a Koopman-autoencoder construction, or a learned norm-preserving one. But a general operator's powers can explode or vanish, so the orbit leaves the region on which the decoder was trained. A norm-preserving operator whose eigenvalue angles are generic avoids explosion but does not return exactly, so phase error accumulates around the loop. Pinning the spectrum to roots of unity makes the operator both norm-preserving and exactly closing. Exact closure is therefore not an incidental convenience: it removes one entire failure mode from the learned representation.

The constraint, rather than the mere presence of a linear operator, carries the construction. Dynamical-systems uses of latent operators usually forecast an open trajectory and have no reason to impose $R^X=\Id$. Here the finite cardinality of the represented collection supplies exactly that reason.

\subsection{Capacity before measurement}

Once closure is guaranteed by $R$, the remaining limitation is representational. The latent requires on the order of $X/2$ rotating planes to resolve $X$ phases, and the decoder must have enough capacity to render the target objects. The number of crisp items should therefore scale with the decoder's budget rather than with an error-accumulation ceiling.

One entanglement must be stated in advance. Beyond $X=2m$ phases, a fixed bank of $m$ planes still produces distinct latents, but those latents crowd onto a smoother curve. Degradation beyond roughly twice the plane count therefore mixes latent crowding with decoder capacity; only the regime up to $2m$ cleanly isolates the decoder.

\section{Relation to prior work}

A fixed linear latent operator under a nonlinear decoder is the Koopman-autoencoder construction of \citet{lusch2018}, and learned norm-preserving operators through orthogonal or unitary parameterizations are established by \citet{arjovsky2016} and \citet{lezcano2019}. Our operator is the special case whose eigenvalues are roots of unity---the discrete Fourier operator---but the use is different. That literature learns operators to forecast one system's open trajectory. We fix the operator to an exact period and use its orbit to represent a finite collection, then account for capacity and compression.

Latent rotations that render transformed views of a single object \citep{worrall2017} and latent spaces with circular topology \citep{falorsi2018} store one object's orbit under a group action, not a collection of distinct images, and provide no capacity or compression accounting of the kind measured here. Cyclic sequence retrieval in Hopfield-style networks \citep{sompolinsky1986} asks a capacity question for binary patterns without a learned continuous rotation or an image decoder.

On the video side, implicit neural representations store signals in weights \citep{sitzmann2020,dupont2021}, and the NeRV line specializes the idea to per-clip video with a frame-index input \citep{chen2021}, refined through hierarchical encodings \citep{kwan2023} and phase-shifted sinusoids \citep{mai2022}. These methods do not make the temporal embedding periodic in the loop length, which is exactly the property isolated by our seam measurement.

Loop synthesis is an active line: latent-shift cycling in video diffusion \citep{mobius2025}, loopable animation training \citep{loopanimate2024}, and 3D cinemagraphs \citep{loopgaussian2024} generate looping content from text or images. Our construction is complementary: it is a fitted representation of a given loop with closure guaranteed by the latent operator, in the lineage of the classical video-textures problem \citep{schodl2000} posed for neural representations.

The combination claimed here---distinct images or frames as phases of one exactly periodic operator, a codebook-controlled capacity and compression account, and a measured closure guarantee---is, to our knowledge, not present in this prior work. The broader finite-knowledge interpretation does not change that narrow novelty claim; it explains what kind of representation the combination instantiates.

\section{Exact recurrence as the representation constraint}
\label{sec:operator}

Holding the decoder fixed and varying only the operator isolates the claim directly. On synthetic $16\times16$ images, reconstruction PSNR in dB is:

\begin{table}[h]
\centering
\small
\begin{tabular}{lrrrr}
\toprule
operator & $X=8$ & $16$ & $32$ & $64$\\
\midrule
general linear (Koopman-style) & 60.5 & 27.0 & 56.5 & 19.0\\
learned unitary (norm-preserving, not periodic) & 56.8 & 66.9 & 69.9 & 59.9\\
\textbf{exact-periodic Fourier (ours)} & \textbf{70.7} & \textbf{73.1} & \textbf{71.5} & \textbf{68.3}\\
learned spectrum (frequencies trained) & 55.4 & 72.7 & 71.3 & 68.6\\
\bottomrule
\end{tabular}
\caption{Operator comparison on synthetic images. The decoder is held fixed across operators.}
\label{tab:synthetic}
\end{table}

The general operator is erratic and collapses at larger $X$; the unitary operator is stable but degrades because its cycle does not close; the exactly periodic operator is best and flat. Learning the frequencies does not help: integer harmonics already form a complete basis. The operative ingredient is exact periodicity itself.

On real images the gap widens, with one concession forced by the measurements. Taking one CIFAR-100 class at $32\times32$ grayscale and the same three operators over a shared decoder gives:

\begin{table}[h]
\centering
\small
\begin{tabular}{lrrrrr}
\toprule
operator & $X=16$ & $64$ & $128$ & $256$ & $500$\\
\midrule
general linear (Koopman-style) & \textbf{82.2} & 61.4 & -- & -- & --\\
learned unitary (not periodic) & 76.4 & 47.5 & 46.5 & 13.5 & 12.2\\
\textbf{exact-periodic Fourier (ours)} & 68.9 & \textbf{71.3} & \textbf{72.6} & \textbf{43.3} & \textbf{52.4}\\
\bottomrule
\end{tabular}
\caption{Operator comparison on real CIFAR-100 images, one class.}
\label{tab:real}
\end{table}

At $X=16$, the unconstrained operator is the best fit by thirteen decibels: with few phases to hold, freedom fits better and the constraint buys nothing. The constraint pays as the orbit grows. Beyond $X=64$, the general operator's iterated powers overflow and no finite reconstruction is obtained (the dashes). The non-periodic unitary operator collapses from 76 to roughly 13 dB as its unclosed cycle accumulates error on real detail. The exactly periodic operator remains near 70 dB through 128 images. Beyond that, as the represented collection approaches the decoder's budget, reconstruction falls into the forties while still leading by roughly thirty decibels. The non-monotonicity of the last two entries, 43.3 then 52.4, is optimization variance at budget saturation rather than a capacity effect. The crisp ceiling of roughly one hundred images is set by the decoder, quantified next.

\begin{figure}[t]
\centering
\includegraphics[width=\textwidth]{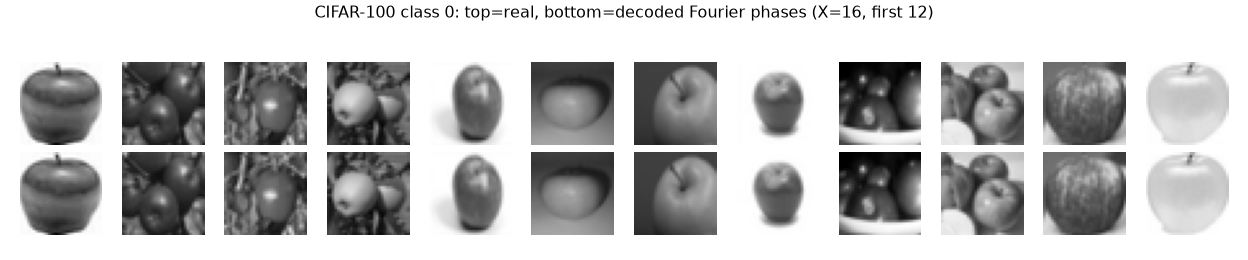}
\vspace{0.4em}
\includegraphics[width=0.78\textwidth]{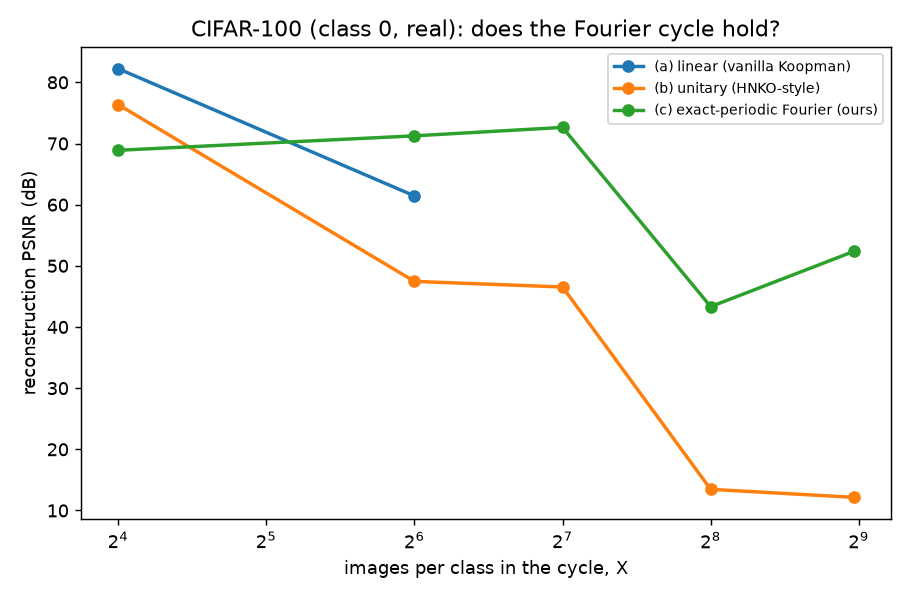}
\caption{CIFAR-100, one class, real images. Top: real images (upper row) and decoded Fourier phases (lower row) at $X=16$; one operator orbit reproduces the class images. Bottom: reconstruction versus the number of images held. The exactly periodic operator stays crisp to roughly one hundred images, while the general linear operator diverges and the non-periodic unitary operator collapses.}
\label{fig:cifar}
\end{figure}

The finite-knowledge reading of these measurements is precise. A cyclic generator can serve as a stable address system for many independently rendered items, but only if the address dynamics close exactly. The more expressive unconstrained operator wins when the collection is small enough that long-run stability is irrelevant; the structured operator wins when the representation must remain valid around a longer orbit.

\section{Capacity and structural sharing}

Three measurements turn the capacity claim into an accounting exercise.

First, there are no free items between trained phases. Training a 32-phase cycle on only sixteen phases leaves the sixteen untrained phases as blurry class-average blobs, forty decibels below the trained ones. The operator interpolates in latent space, but the decoder renders only what it was fit to render. Adding a denoising objective---rendering from noised latents, the primitive beneath diffusion models---does not change the result: at every noise level tested, the untrained phases improve by at most 0.2 dB while the trained phases lose up to 2.7 dB. The midpoint between unrelated images corresponds to no image, and local smoothing does not invent one. A cyclic indexing of arbitrary finite items therefore supplies order and closure, not semantic interpolation. Intermediate phases become meaningful only when the content itself supplies a manifold, as in the video experiment of Section~\ref{sec:video}.

Second, the crisp ceiling tracks decoder width. Across dense decoders from thirty-two thousand to 2.2 million parameters, the number of images stored crisply scales with hidden width $h$, while parameter count scales as approximately $1{,}150h$. This is roughly 1,150 dense parameters per crisp image, which in half precision is about $2.25\times$ the bytes of the raw $32\times32$ grayscale image itself. The factor is structural: for a dense decoder, the break-even set size always lies above the crisp ceiling. At every size tested, the construction can therefore be crisp or compressed, but not both. Extending $X$ to the full fifty-thousand-image corpus makes the trade sharper: compression crosses one only beyond four thousand images, by which point reconstruction has fallen below 20 dB.

Third, weight sharing changes the constant. A small convolutional decoder on 128 images from a single class---objects that share a low-dimensional visual manifold---reaches 32.2 dB, above the 30 dB crisp line, with thirty-nine thousand parameters. At half precision the representation is $1.66\times$ smaller than the raw images: crisp and compressed simultaneously, a regime no dense decoder reached. On 128 maximally diverse images, the same decoder falls just short at 29.3 dB with the same size. The crossing therefore requires both shared kernels and shared structure in the represented collection. The accounting cuts both ways: an orbit representation stores finite objects below raw-instance cost only when the collection contains structure that a shared decoder can exploit. This is not a claim against general-purpose image codecs.

\begin{figure}[t]
\centering
\includegraphics[width=\textwidth]{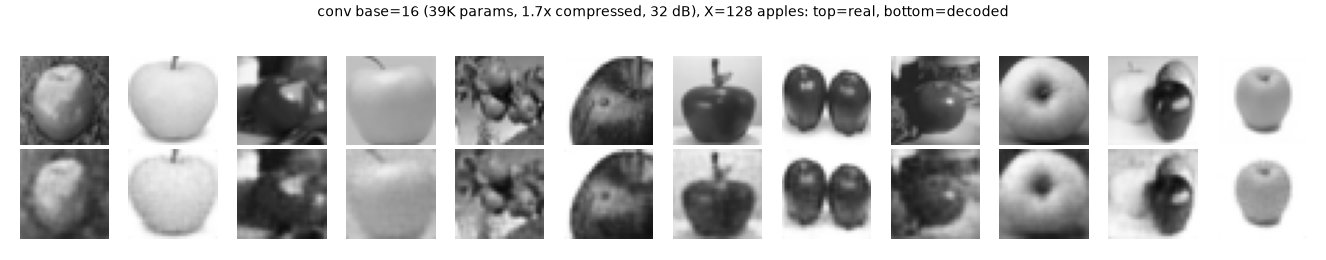}
\caption{The compression crossing. A 39K-parameter convolutional decoder holds 128 single-class images (top: real; bottom: decoded phases) at 32.2 dB while occupying $1.66\times$ fewer bytes than the raw pixels.}
\label{fig:crossing}
\end{figure}

From the finite-knowledge perspective, compression is a consequence rather than the primary object. The generator removes the independent latent table; the decoder compresses whatever regularity is shared across items. If the collection has no shared structure, the decoder must effectively memorize the instances and the economy disappears.

\section{The explicit-storage control: the generative law is free}

The cyclic representation could be buying order and closure at the cost of reconstruction quality. The decisive control replaces the rotating base vector with an independently learned latent for every image: a free codebook with sixteen thousand latent parameters against the cycle's one hundred twenty-eight, evaluated over the same decoders and image sets.

\begin{table}[h]
\centering
\small
\begin{tabular}{llrrrr}
\toprule
setting & decoder & \textbf{cycle (ours)} & codebook & gap & compression (cycle / codebook)\\
\midrule
single-class & base 16 & \textbf{32.4} & 32.4 & $+0.0$ & $1.66\times / 1.18\times$\\
single-class & base 32 & \textbf{39.6} & 38.1 & $-1.5$ & $0.72\times / 0.61\times$\\
diverse & base 16 & \textbf{29.4} & 29.0 & $-0.4$ & $1.66\times / 1.18\times$\\
diverse & base 32 & \textbf{37.4} & 36.0 & $-1.4$ & $0.72\times / 0.61\times$\\
\bottomrule
\end{tabular}
\caption{Cycle versus one independent learned latent per image. Negative gap means the cycle reconstructs better.}
\label{tab:codebook}
\end{table}

The free latents never win. They tie at the smaller decoder and lose modestly at the larger decoder on both the shared-manifold and diverse sets, while requiring a latent store one hundred twenty-eight times larger and producing a visibly worse compression ratio. The pattern holds over three retraining seeds with fresh image selections: the codebook's mean advantage never exceeds $+0.14$ dB in any cell, with a best single run of $+0.47$ dB. At the larger decoder, the cycle leads by more than a decibel in both settings, with gaps of $-1.16\pm0.20$ and $-1.33\pm0.25$.

The structural constraint therefore costs nothing in reconstruction on these studies. It buys parameter economy and three properties the independent codebook cannot provide by itself: a total ordering, a one-step transition from every item to the next, and exact closure. In extension-versus-intension terms, the codebook stores the collection as instances; the cycle stores it as a generated orbit.

\section{When phase is meaning: looping video}
\label{sec:video}

For a looping clip, the cyclic order is not an arbitrary address assignment but the content's own temporal structure. The construction becomes a per-clip video representation in the NeRV family: a network fit to one clip, rendering frame $t$ from an embedding of $t$, with the clip stored in the network weights. The distinguishing choice is the embedding. NeRV-style models embed the frame index through sinusoids at geometric frequencies that are not periodic in the loop length. We use the exactly periodic cycle---the rotating base vector of Section~2---so frame $T$ equals frame 0 to machine precision by construction rather than by training.

We fit palindromed loops, in which a clip is played forward and then backward so that every benchmark clip closes exactly. Seven standard sources are used: the Big Buck Bunny excerpt used by NeRV, the Bunny and Sintel Blender trailers, UVG HoneyBee and Jockey as near-static and fast-motion poles \citep{mercat2020}, and DAVIS blackswan and camel \citep{ponttuset2017}. All clips are represented at $128\times128$ RGB with one decoder architecture across encoders. We measure fidelity, representation size relative to raw frames, and the loop seam, defined as the mean absolute difference between the rendered frame at the wrap point and at phase zero.

\begin{table}[h]
\centering
\small
\begin{tabular}{lccc}
\toprule
clip & \textbf{cycle (ours)} & frame index & harmonic index\\
\midrule
Big Buck Bunny (NeRV clip) & {\boldmath$33.3\pm0.2$} & $32.8\pm0.2$ & $32.7\pm0.1$\\
Bunny trailer & {\boldmath$38.5\pm0.9$} & $38.1\pm2.2$ & $37.4\pm0.1$\\
Sintel trailer & {\boldmath$38.3\pm0.5$} & $36.9\pm0.4$ & $29.5\pm6.2$\\
UVG HoneyBee & {\boldmath$36.0\pm0.2$} & $35.9\pm0.4$ & $35.3\pm0.5$\\
UVG Jockey & {\boldmath$30.1\pm0.3$} & $28.9\pm0.2$ & $29.2\pm0.2$\\
DAVIS blackswan & {\boldmath$26.9\pm0.2$} & $26.3\pm0.0$ & $26.4\pm0.1$\\
DAVIS camel & {\boldmath$24.9\pm0.0$} & $24.1\pm0.1$ & $24.1\pm0.1$\\
\bottomrule
\end{tabular}
\caption{Looping-video PSNR in dB, mean and standard deviation over three retraining seeds, at equal parameter count.}
\label{tab:video}
\end{table}

Every run is roughly $2.5\times$ smaller than the raw frames. The cycle leads the frame-index baseline on all seven clips, by 0.7 dB on average. On the seam, the two representations differ by five orders of magnitude on every clip and seed: the cycle wraps at numerical precision, never exceeding $3.3\times10^{-7}$ mean absolute intensity, while the frame-index representation never falls below $1.5\times10^{-2}$, producing a visible pop. The seam makes the representation claim measurable. Exact closure is a property that the standard embedding does not have and cannot reliably learn; the cycle has it by construction.

The video setting also reverses the earlier no-free-items result, with sampling theory specifying the condition. Training on every other frame and evaluating the held-out frames fails badly with the full harmonic bank, at 10--18 dB, below even copying the nearest trained frame. The reason is aliasing: when only half the phases are observed, frequencies above the training grid's Nyquist limit are unconstrained and oscillate freely between samples. Capping the bank at Nyquist repairs the failure. Held-out fidelity rises by nine to twenty-two decibels across three seeds, reaches the model's own trained-frame ceiling on near-static HoneyBee content (35.5--36.0 versus 35.7--36.2 dB), beats the frame-index embedding on all three clips tested, and matches or beats frame copying on the dynamic clips. The bandlimited cycle is the Whittaker--Shannon interpolator on the circle, so a loop represented at one frame rate can be rendered at another from the representation alone. Unrelated images provide no manifold between phases; consecutive video frames do, provided that the spectrum respects the sampling grid.

A fair alternative is to pin the baseline's frequencies directly to integer harmonics of the loop. We run that control. The harmonically pinned embedding closes its seam at or below $8.5\times10^{-8}$ everywhere, confirming that exact periodicity, however imposed, is the ingredient responsible for closure. It does not match the cycle's fidelity: it trails by 1.9 dB on average, is behind on all seven clips, and is unstable on one clip, the Sintel entry with a 6.2 dB spread caused by one collapsed run. The constraint is necessary, and the carrier matters. Routing the harmonics through a learned embedding network costs fidelity and stability relative to the fixed rotation of a single base vector, which provides the same harmonic structure with zero additional parameters.

Two boundaries complete the result. Compression is measured against raw frames and a matched baseline at equal training, as appropriate for a per-clip neural representation; conventional video codecs compress these clips much further and are not the comparison. The palindrome protocol makes every clip exactly loopable, so the seam measures the representation rather than a mismatch in the content. For natively non-looping content played once, the construction offers no advantage over a standard frame-index embedding.

\section{What the finite-knowledge perspective adds}

The measurements support a representation principle narrower and more defensible than the claim that arbitrary databases should become dynamical systems. A finite collection can be encoded as an orbit when three roles are separated:

\begin{enumerate}[leftmargin=1.5em,itemsep=0.2em]
\item the group action supplies addresses, order, and exact recurrence;
\item the decoder supplies the content associated with each address;
\item shared structure in the content determines whether the representation compresses or interpolates.
\end{enumerate}

For unrelated objects, the orbit is a compact addressing law. It does not create semantic neighborhoods between adjacent phases, as the missing-phase experiment demonstrates. For naturally cyclic collections, the orbit can also encode genuine transition structure. Looping video is the measured example; periodic simulations, robotic gaits, biological cycles, and procedural animation are plausible extensions precisely because their finite states already possess a cyclic order and require recurrence. These are applications of the same representation, not empirical claims established here.

The paper also exposes the limit of the cyclic group. Taxonomies, graphs, permutation spaces, and branching workflows are not naturally one-dimensional cycles. Forcing them into one orbit would provide an enumeration but could destroy their native adjacency. A broader program would replace the single cyclic generator with one or more generators of a finite group or semigroup acting on latent space, so that the algebra of the representation matches the combinatorial structure of the knowledge. The present study establishes only the $C_X$ case: one exactly periodic generator, one closed orbit, and one shared decoder.

This framing clarifies why compression is secondary. The codebook control shows that the central economy is not necessarily in the decoder weights; it is in replacing $X$ independent latent entries with a law. Whether the full representation is smaller than the raw objects then depends on the decoder's ability to exploit shared structure. In that sense, the work is about compressing relations---the regularity of how items are addressed and rendered---rather than merely compressing pixels.

\section{Scope}

Everything reported here is small-scale and internally controlled: grayscale $32\times32$ images, $128\times128$ clips, decoders from tens of thousands to a few million parameters, and every comparison at matched decoder and training conditions. The codebook control and video study are replicated over three retraining seeds. The operator-comparison tables in Section~\ref{sec:operator} remain single-seed, with margins of tens of decibels. All experiments were designed, run, and verified on Apple silicon (an M1 and an M4 machine); no cloud compute was used.

The capacity ceiling is the decoder budget. Compression claims are against raw pixels and matched baselines, never against image or video codecs. Achieving both crisp reconstruction and compression requires shared structure in the represented collection. An imposed cyclic order provides deterministic addressing and closure but does not make unobserved phases meaningful for unrelated objects. On content that does not loop, exact periodicity offers no advantage. The present operator represents a cyclically indexed finite collection; it is not yet a representation of arbitrary graphs, taxonomies, or branching knowledge structures.

Within these limits, the claims are three. First, exact periodicity is the structural constraint that lets one operator enumerate many objects without accumulated drift. Second, the constraint costs nothing against independent latent codes in the measured reconstruction settings while using far less latent storage and supplying order and exact closure. Third, when the represented knowledge is naturally cyclic, as in looping video, the construction converts an approximate training outcome---the seam---into a property guaranteed by the representation.

\section*{Acknowledgements}
All experiments in this paper were designed, run, and verified on Apple silicon (an M1 and an M4 machine), and no cloud compute was used. We thank Aarav Pal for his help with the data downloads and with testing and running the experiments.


\begin{thebibliography}{99}

\bibitem[Arjovsky et~al.(2016)Arjovsky, Shah, and Bengio]{arjovsky2016}
M. Arjovsky, A. Shah, and Y. Bengio.
\newblock Unitary Evolution Recurrent Neural Networks.
\newblock \emph{ICML}, 2016. arXiv:1511.06464.

\bibitem[Chen et~al.(2021)Chen, He, Wang, Ren, Lim, and Shrivastava]{chen2021}
H. Chen, B. He, H. Wang, Y. Ren, S.-N. Lim, and A. Shrivastava.
\newblock NeRV: Neural Representations for Videos.
\newblock \emph{NeurIPS}, 2021. arXiv:2110.13903.

\bibitem[Dupont et~al.(2021)Dupont, Goli\'nski, Alizadeh, Teh, and Doucet]{dupont2021}
E. Dupont, A. Goli\'nski, M. Alizadeh, Y. W. Teh, and A. Doucet.
\newblock COIN: COmpression with Implicit Neural Representations.
\newblock 2021. arXiv:2103.03123.

\bibitem[Falorsi et~al.(2018)Falorsi, de Haan, Davidson, De Cao, Weiler, Forr\'e, and Cohen]{falorsi2018}
L. Falorsi, P. de Haan, T. R. Davidson, N. De Cao, M. Weiler, P. Forr\'e, and T. S. Cohen.
\newblock Explorations in Homeomorphic Variational Auto-Encoding.
\newblock 2018. arXiv:1807.04689.

\bibitem[Kwan et~al.(2023)Kwan, Gao, Zhang, Gower, and Bull]{kwan2023}
H. M. Kwan, G. Gao, F. Zhang, A. Gower, and D. Bull.
\newblock HiNeRV: Video Compression with Hierarchical Encoding-based Neural Representation.
\newblock \emph{NeurIPS}, 2023. arXiv:2306.09818.

\bibitem[Lezcano-Casado and Mart\'inez-Rubio(2019)]{lezcano2019}
M. Lezcano-Casado and D. Mart\'inez-Rubio.
\newblock Cheap Orthogonal Constraints in Neural Networks.
\newblock \emph{ICML}, 2019. arXiv:1901.08428.

\bibitem[LoopAnimate(2024)]{loopanimate2024}
LoopAnimate: Loopable Salient Object Animation.
\newblock 2024. arXiv:2404.09172.

\bibitem[LoopGaussian(2024)]{loopgaussian2024}
LoopGaussian: Creating 3D Cinemagraph with Multi-view Images via Eulerian Motion Field.
\newblock 2024. arXiv:2404.08966.

\bibitem[Lusch et~al.(2018)Lusch, Kutz, and Brunton]{lusch2018}
B. Lusch, J. N. Kutz, and S. L. Brunton.
\newblock Deep Learning for Universal Linear Embeddings of Nonlinear Dynamics.
\newblock \emph{Nature Communications}, 9:4950, 2018.

\bibitem[Mai and Liu(2022)]{mai2022}
L. Mai and F. Liu.
\newblock Motion-Adjustable Neural Implicit Video Representation.
\newblock \emph{CVPR}, 2022.

\bibitem[Mercat et~al.(2020)Mercat, Viitanen, and Vanne]{mercat2020}
A. Mercat, M. Viitanen, and J. Vanne.
\newblock UVG Dataset: 50/120fps 4K Sequences for Video Codec Analysis and Development.
\newblock \emph{ACM MMSys}, 2020.

\bibitem[Mobius(2025)]{mobius2025}
Mobius: Text to Seamless Looping Video Generation via Latent Shift.
\newblock 2025. arXiv:2502.20307.

\bibitem[Pont-Tuset et~al.(2017)Pont-Tuset, Perazzi, Caelles, Arbel\'aez, Sorkine-Hornung, and Van Gool]{ponttuset2017}
J. Pont-Tuset, F. Perazzi, S. Caelles, P. Arbel\'aez, A. Sorkine-Hornung, and L. Van Gool.
\newblock The 2017 DAVIS Challenge on Video Object Segmentation.
\newblock 2017. arXiv:1704.00675.

\bibitem[Sch\"odl et~al.(2000)Sch\"odl, Szeliski, Salesin, and Essa]{schodl2000}
A. Sch\"odl, R. Szeliski, D. H. Salesin, and I. Essa.
\newblock Video Textures.
\newblock \emph{SIGGRAPH}, 2000.

\bibitem[Sitzmann et~al.(2020)Sitzmann, Martel, Bergman, Lindell, and Wetzstein]{sitzmann2020}
V. Sitzmann, J. N. P. Martel, A. W. Bergman, D. B. Lindell, and G. Wetzstein.
\newblock Implicit Neural Representations with Periodic Activation Functions (SIREN).
\newblock \emph{NeurIPS}, 2020. arXiv:2006.09661.

\bibitem[Sompolinsky and Kanter(1986)]{sompolinsky1986}
H. Sompolinsky and I. Kanter.
\newblock Temporal Association in Asymmetric Neural Networks.
\newblock \emph{Physical Review Letters}, 57(22):2861, 1986.

\bibitem[Worrall et~al.(2017)Worrall, Garbin, Turmukhambetov, and Brostow]{worrall2017}
D. E. Worrall, J. S. Garbin, D. Turmukhambetov, and G. J. Brostow.
\newblock Interpretable Transformations with Encoder-Decoder Networks.
\newblock \emph{ICCV}, 2017. arXiv:1710.07307.

\end{thebibliography}
\end{document}